\documentclass[conference]{IEEEtran}
\usepackage{blindtext, graphicx}
\ifCLASSINFOpdf
\else
\fi
\usepackage{hyperref}

\begin{document}
%
\title{Massively Deep Artificial Neural Networks for Handwritten Digit Recognition}

\author{\IEEEauthorblockN{Keiron-Teilo O'Shea}
\IEEEauthorblockA{Department of Computer Science\\
Aberystwyth University\\
Aberystwyth, Ceredigion, SY23 3DB\\
keo7@aber.ac.uk
}}


%


\maketitle

\begin{abstract}
Greedy Restrictive Boltzmann Machines yield an fairly low 0.72\% error rate on the famous MNIST database of handwritten digits. All that was required to achieve this result was a high number of hidden layers consisting of many neurons, and a graphics card to greatly speed up the rate of learning.
\end{abstract}

\begin{IEEEkeywords}
ANN (Artificial Neural Networks), RBM (Restrictive Boltzmann Machine), MNIST  handwritten database\footnote{\url{http://yann.lecun.com/exdb/mnist/}}, GPU (Graphics Processing Unit)
\end{IEEEkeywords}

\section{Introduction}

The task of recognising handwritten digits is of great interest for both academic and commercial application. \cite{miramontes2011assessment} Existing contemporary algorithms are already adept at learning to recognise handwritten digits, and are being used within Post Offices for automated letter sorting. The MNIST database of handwritten digits is understood to be the most popular benchmark for this form of pattern recognition task. \cite{lecun-mnisthandwrittendigit-2010}

Some years ago, a class of artificial neural networks called Restrictive Boltzmann Machines (RBMs) \cite{Smolensky:1986:IPD:104279.104290} were amongst one of the first the initial classifiers tested on the MNIST data set \cite{tang2011data}. RBMs are a variant of standard Boltzmann machines, but with the restriction that their neurons must form a bipartite graph; where a pair are ``visible" and ``hidden" units are used respectively.

Restricted Boltzmann machines are commonly used to formulate deep neural networks. Deep belief networks can be formed by ``stacking'' RBM's, and finetuning the network using the gradient descent optimisation algorithm and Backpropagation. \cite{lecun1989backpropagation} \cite{hinton2006fast}

The first documented use of Convolutional Neural Networks (CNN) \cite{simard2003best} achieved a world-record 0.40\% error rate when given the task of classifying MNIST digits. Recently, better results have been obtained by pre-training each hidden CNN layer one by one in an unsupervised manner achieving an incredibly low error rate of 0.39\%. \cite{boureau2008sparse}

The downside of using these CNNs are that they are both extremely resource heavy and time consuming. Online backpropagation for thousands of epochs on large RBMs could take months on even the newest standard off-the-shelf desktop microprocessors. One option could be to parallelise the workload across a computing cluster, but latency issues between individual computers may prove difficult to overcome. Multi-threading on a multi-core processor is difficult on a dataset the size of MNIST, as it's too large to fit in the L2/L3 cache of most desktop microprocessors. So to train large RBMs, it will be required to continually access data from RAM; which only causes further latency issues. \cite{jang2008neural}

However, as desktop Graphics Processing Units (GPUs) have become faster, the large amount of on-board memory allows the possibility of training large RBMs quickly and efficiently. \cite{chellapilla2006high}

\section{Data}

The MNIST database contains 60,000 digits from 0 to 9. Some examples are shown below in Figure \ref{fig:mnist}. The standard MNIST dataset is comprised of two sets, one for training (50,000 images), and one for testing (10,000 images). It is common to split the data set into two sets, 50,000 images are used for training, where a further 10,000 images are kept for validation. Our network is trained on standard MNIST digits. Pixel intensities of the standard MNIST data set range from 0 (being the white background) and to 255 (complete black). $28 \times 28 = 784$ pixels per MNIST images, mapped to real values ${pixel intensity \over 127.5} - 1.0$ in [-1.0, 1.0] are fed into the input layer of the Artificial Neural Network.

\begin{figure}[h]
\centering
\includegraphics[width=0.45\textwidth]{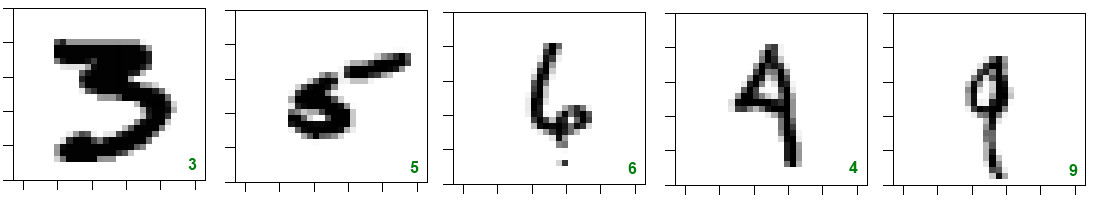}
\caption{Examples of MNIST data set}
\label{fig:mnist}
\end{figure}

\section{Network Architecture}

Training was done on a simplistic Restrictive Boltzmann Machine (RBM) containing 2 to 9 hidden layers, and shifting numbers of hidden units. The number of hidden units per layer typically shrank toward the output layer. (Table \ref{tab:testresults})

On-line Backpropagation was used, without the use of momentum or DropOut. The learning rate was set to vary on each epoch, starting from $10^{-3}$ leading downwards to $10^{-6}$. Weights are initially set to a uniform random distribution in [-0.05, 0.05], and a decay of the weights being set at to 0.01. Each neuron's activation function was set as an scaled hyperbolic tangent:

\begin{equation}
y\left(a\right)\space =\space A\space \tanh \left(Ba\right)
\end{equation}

Where $A$ = 1.71 and $B$ = 0.66. The binary visible units were set to independent Gaussian, where rectified hidden units were used to further the expression capabilities of all hidden neurons.

\section{Results}

All tests were ran on a computer with a Intel i5-2500k 3.0GHz processor, 16GB of DDR3 RAM, and a nVidia GTX 580 graphics card with 3GB of GDDR5 memory. The GPU was used to accelerate the performance of both the forward propagation and backpropagation routines. The trained RBM with the lowest validation error was selected, and then used to evaluate the performance on the MNIST test set. Results are summarised in Table \ref{tab:testresults}.

The best neural network has an error rate of only 0.72\% (72 out of 10,000 incorrectly classified). Investigation has proved that the majority of the 34 misclassified digits feature few or no main attributes, meaning that even human perception will find difficult to correctly identify.

The best test error of this particular RBM was even lower (0.40\%), and has been identified as the maximum capacity of the network. It is obvious that performance increases greatly by adding hidden layers and more units per hidden layer. Example being that the network 5 in Table \ref{tab:testresults}

Networks that contain up to 12 million weights can be trained using the standard gradient descent algorithm to achieve test errors below the 2\% mark after 30-45 epochs in less than 3 hours of training.

\begin{table}[h]
\centering
\begin{tabular}{cccc}
\multicolumn{1}{c}{\begin{tabular}[c]{@{}c@{}}architecture\\ (\# hidden neurons)\end{tabular}} & \multicolumn{1}{c}{\begin{tabular}[c]{@{}c@{}}test error [\%] \\ best evaluation [\%] \end{tabular}} & \multicolumn{1}{c}{\begin{tabular}[c]{@{}c@{}}best test \\ error\end{tabular}} & \multicolumn{1}{c}{time {[}hours{]}} \\ \hline
1000, 500                                                                                                      & 0.92                                                                                       & 0.90                                                                           & 16.3                             \\
1500, 1000, 500                                                                                                & 0.85                                                                                      & 0.83                                                                            & 26.3                             \\
2500, 1500, 100, 500                                                                                           & 0.78                                                                                       & 0.76                                                                           & 45.2                             \\
2500, 2000, 1500, 1000, 750                                                                                    & 0.72                                                                                      & 0.7                                                                            & 83.1                             \\ \hline
9 x 1000                                                                                                       & 0.88                                                                                      & 0.85                                                                            & 77.3                            
\end{tabular}
\label{tab:testresults}
\caption{Error rates on MNIST test set}
\end{table}

\section{Conclusion}

As computing power becomes more affordable, it will greatly push forward the boundaries of machine learning techniques. Modern-day GPUs are already more than 20 times faster than standard general purpose multiprocessors when faced with the task of training big and deep artificial neural networks.

On an extremely difficult MNIST handwritten benchmark, the use of standard off the shelf GPU-based neural networks have surpassed all previously reported results, including all scores obtained using complex specialised architectures. Of course, this approach is not limited to the task of classifying handwritten digits, and holds great promise for all pattern recognition tasks.


\section*{Acknowledgment}

This work begun during the course of the authors undergraduate dissertation. He would like to thank his project supervisor, Chuan Lu, for her guidance, and Adam Gibson for providing the deeplearning4j deep learning framework.



%

\nocite(*)
\bibliographystyle{abbrv}
\bibliography{bib}

\section*{Appendices}

\subsection{GPUs and Artificial Neural Networks}

Previously, the only way to program a GPU was to create a set of graphical operations using technologies such as DirectX and OpenGL. Despite these limitations, people were still able to hard code and implement a number of GPU-based Artificial Neural Networks. 

Due to the added complexity, these networks were typically shallow. But a noticeable, if not modest speedup was observed with the use of GPUs. 2007 saw NVIDIA announce their first foray into scientific computing with CUDA (the Compute Unified Device Architecture), a C-like programming language for scientific use. GPUs have a greater amount of pure processing speed and memory bandwidth, when compared to most microprocessors; and this allows for quick and effective ANN implementations.

%
\IEEEpeerreviewmaketitle

\end{document}